\documentclass[runningheads]{llncs}
\usepackage{graphicx}
\usepackage{amssymb}
\usepackage{mathabx}
\usepackage{subcaption}
\usepackage{tikz}
\usepackage{pgfplots} 
\usepackage{xcolor}

\begin{document}
\title{ConvNeXtv2 Fusion with Mask R-CNN for Automatic Region Based Coronary Artery Stenosis Detection for Disease Diagnosis}
%
%
\author{Sandesh Pokhrel\thanks{Equal contribution}\inst{1}\orcidID{0009-0001-4843-7899}\, 
Sanjay Bhandari\textsuperscript{*} \inst{1}\orcidID{0009-0009-0722-0739}\and
Eduard Vazquez \inst{2} \and
Yash Raj Shrestha \inst{3} and 
Binod Bhattarai \inst{4}\orcidID{0000-0001-7171-6469}}
\authorrunning{Pokhrel \& Bhandari et al.}
%
\institute{Nepal Applied Mathematics and Informatics Institute for research(NAAMII), Lalitpur, Nepal \and
Fogsphere(Redev AI Ltd.), 64 Southwark Bridge Rd, SE1 0AS, London, UK \and
University of Lausanne, Switzerland \and 
School of Natural and Computing Sciences, University of Aberdeen, Aberdeen, UK\\
}
\maketitle              
\begin{abstract}

Coronary Artery Diseases although preventable are one of the leading cause of mortality worldwide. Due to the onerous nature of diagnosis, tackling CADs has proved challenging. This study addresses the automation of resource-intensive and time-consuming process of manually detecting stenotic lesions in coronary arteries in X-ray coronary angiography images. To overcome this challenge, we employ a specialized Convnext-V2 backbone based Mask RCNN model pre-trained for instance segmentation tasks. Our empirical findings affirm that the proposed model exhibits commendable performance in identifying stenotic lesions. Notably, our approach achieves a substantial F1 score of 0.5353 in this demanding task, underscoring its effectiveness in streamlining this intensive process.

\keywords{Coronary Artery Segmentation \and Stenosis \and CADs \and Instance Segmentation \and Mask R-CNN \and ConvNeXt-V2}
\end{abstract}

\section{Introduction}
Coronary Artery Disease(CAD) is a medical condition arising due to the restriction of blood flow in the coronary arteries caused by the accumulation of atherosclerotic plaque in the coronary arteries. 
CADs are the third leading cause of mortality worldwide and are associated  with 17.8 million deaths annually~\cite{riskfactor}. Even though it is a significant cause of death and disability, it is preventable through proper diagnosis.
The primary diagnostic approach for CAD is coronary angiography, which involves the application of contrast agent in the arterial region to detect any leisons in the artery segments that can be analyzed through X-ray images by the physicians.
Through careful analysis of the vessels and arterial regions in the images physicians determine the extent of blockage and the severity of segments affected following up to revascularization procedures if necessary. This direct method of analysis of angiographic images and videos is greatly influenced by physicians' experience which lacks accuracy, objectivity and consistency~\cite{accuracy}. Automated detection and segmentation of arterial regions attempts to help reduce these diagnostic inaccuracies leading to faster, more accurate and more consistent diagnosis.

Invasive X-ray angiography disease diagnosis has been a topic of research for a recent years with the application  of deep learning and neural networks. Coronary Artery detection and segmentation have been facilited through 
Unets~\cite{localcontexttransformer,autostendiag}, 
DenseNet~\cite{densenet},
and 3DCNNs~\cite{accuracy,centerlineExtraction}. %

MAE(Masked Autoencoders)~\cite{mae} are currently the best in the field of vision learning with superior performance in detection and segmentation tasks compared to other self supervised models. They achieve such result by masking random patches of the input image at encoder side and reconstructing the missing pixels at decoder side. With the motive of leveraging this advantage of masked autoencoders in medical domain, we propose to use Convnext-V2~\cite{Convnextv2} backbone based Mask R-CNN~\cite{maskrcnn} architecture for stenosis detection. ConvNeXt-V2 was opted as the backbone as it further enhances the performance of its predecessor ConvNeXt\cite{convnext} through the integration of the Global Response Normalization (GRN) layer. This addition serves to diminish the occurrence of redundant activations, while simultaneously amplifying feature diversity during training. GRN is applied on high-dimensional features within each block, thereby contributing to the model's improved capabilities.

By combining masked autoencoders (MAE) and the Global Response Normalization (GRN) layer, ConvNeXt V2 achieves superior performance in various downstream tasks.With the promise of rewarding results in 
COCO instance segmentation with higher level of adaptability and effectiveness, we found it suitable to be used as backbone in Mask R-CNN based model in the stenosis detection task.

\section{Literature Review}

Coronary Artery Disease being the third leading cause of death and disability~\cite{riskfactor} has been a growing topic of research in medical imaging. The diagnosis of CADs can be done in either a non-invasive way~\cite{noninvasiveECG,SPECT-MPI} or in an invasive manner. Coronary angiogram which is a invasive method of diagnosis is well known as the "gold standard"~\cite{goldstandard} in CAD diagnosis. In an attempt to automate detection of CADs, non-invasive deep learning methods are facilitating the analysis of ECG~\cite{noninvasiveECG,ecgstenosis} and SPECT-MPI~\cite{SPECT-MPI} signals. There are also a number of decision support systems~\cite{aidss,mldss,stendss} which provide assistance to the experts in the diagnosis of these diseases. But being non-invasive they lack diagnosis accuracy of invasive coronary angiography method.

Even though coronary angiograpy method is the most reliable method of diagnosis for CADs, there is still potential for improvement in consistency and accuracy of the diagnosis ~\cite{accuracy}. As a result, deep learning approaches to analyzing angiographic images through segmentation of coronary arteries have emerged as suitable tools for clinicians. Exploring the deep learning approach for segmentation and stenosis detection, Cervantes-Sanchez, et.al. ~\cite{automaticMultiScale} proposed automatic segmentation of coronary arteries in X-ray angiograms based on multi-scale Gabor and Gaussian filters along with multi layer perceptrons. Some works have focused on the view angle of the angiographic image as the segments visible in X-ray images are dependent on view angle. For example,  a two-step deep-learning framework to partially automate the detection of stenosis from X-ray coronary angiography images~\cite{automatedsten} includes automatically identifying and classifying the angle of view and  then determining the bounding boxes of the regions of interest in frames where stenosis is visible. AngioNet~\cite{angionet}, which uses the Angiographic Processing Network combined with Deeplabv3~\cite{deeplabv3} has the design to facilitate detections under poor contrast and lack of clear vessel boundaries in angiographic images. Automatic CAD diagnosis has also been tackled as a combination of subtasks. First the task of segmentation which extracts region of interest(ROI) from original images followed by identification of sections~\cite{autostendiag}. Utilizing the effectiveness of Mask R-CNN~\cite{maskrcnn} in medical domain, Fu et.al.~\cite{maskrcnnCCTA} proposed using it for segmentation which showed promising results on fine and tubular structures of the coronary arteries.

Instead of single images, methods have also been devised to work with consecutive frames using a 3D convolutional network to segment the coronary artery~\cite{accuracy}.
Recurrent CNNs in conjunction  with 3D convolutions have proved helpful in automatic detection and classification of Coronary Artery Plaque and Stenosis in Coronary Angiography. A 3D convolutional neural network~\cite{recurrentcnn} is utilized to extract features along the coronary artery while the extracted features from a recurrent neural network are aggregated to perform two simultaneous multi-class classification tasks. Exploring further into 3D convolutions, 3D Unet for coronary artery lumen segmentation~\cite{3dunetlumen}, focuses on segmentation of CTCA images for data both with and without detecting the centerline.

Apart from these general trends, Graph Convolutional Networks (GCNs) have been explored to predict vertex positions in a tubular surface mesh for coronary artery lumen segmentation in CT angiography~\cite{GCNsegmentation}. U-nets~~\cite{unet-orig} have been extensively investigated in medical image segmentation and angiographic segmentation. BRU-Net~\cite{Brunet}, a variant of U-Net where bottleneck residual blocks are used instead of internal encoder-decoder components of traditional U-Net~\cite{traditionalunet}, effectively optimizes the use of parameters in the network, making it lightweight and very efficient to work with in X-ray angiography. Further, Sait et.al ~\cite{yolov7andunet} suggests using YOLOv7 as feature extractor followed by hyperparameter tuned UNet++ model~\cite{Unet++}.

Talking of YOLO models~\cite{yoloorig}, they have been emerging in medical image analysis due to their real time inference capability and versatility in object detection. A comprehensive analysis of various YOLO algorithms that were explored in medical imaging from 2018, shows the improving trend of the newer versions of YOLO in their capability as feature extractor as well as in downstream tasks due to their specialized heads~\cite{Qureshi2023}.

More recently, the performance of downstream tasks such as detection and segmentation have been improved with the use of backbones trained in a self-supervised manner~\cite{simclr,colorization,mae,wav2vec2}. These architectures perform even better on specialized tasks when trained on vast amount of unlabeled data before finetuning~\cite{mae,Convnextv2,wav2vec2}. SSL methods can be beneficial in the medical domain as they can leverage large-scale, unannotated image datasets to pre-train models on tasks like predicting rotations~\cite{unsupervisedrotate,ganrotate}, color~\cite{colorization}, or missing pixels~\cite{mae,Convnextv2} in an image. 
One of the emerging backbones in this line of research is ConvNeXt\cite{convnext}. To understand the local and global pathological semantics from neural networks and tackle the class-imbalance problem, BCU-net~\cite{bcunet} leveraged ConvNeXt~\cite{convnext} in global interaction and U-Net~\cite{unet-orig} in local processing on binary classification tasks in medical imaging. ConvNeXt has further proven its effectiveness in medical image segmentation tasks as a backbone capable of improving the performance along with significant reduction in the number of parameters of classical Unet~\cite{unet-orig,convnextskin}. Specifically in tasks relating to arterial segmentation, Convnext has been used to improve classification of RCA angiograms utilizing LCA information~\cite{convnextangiogram}.

The latest iteration ConvNeXtv2~\cite{Convnextv2}, which has improved performance over ConvNeXt due to architectural enhancements, is much less explored in medical imaging tasks and even less so in angiographic images. Inspired by the effectiveness of ConvNeXt in medical imaging as a backbone architecture and considering the implications of improved ConvNextV2, we petition it as a backbone for the task of stenosis detection model.

\section{Methods}

\subsection{Model Pipeline}

We introduce an innovative approach for stenosis detection, leveraging the powerful Mask R-CNN framework. We advocate the use of the Convnext-V2 backbone, enabling the extraction of more enriched and semantically significant feature maps. A Region Proposal Network (RPN) was incorporated to efficiently identify a multitude of potential Regions of Interests (ROIs) prior to the segmentation phase. To tackle the challenge of inconsistent ROI sizes, we implemented ROI-Alignment, facilitating the direct extraction of features from the maps generated by the backbone. 
Furthermore, diverse data augmentation techniques were employed to accommodate for the poor contrast and illumination of the training dataset. In order to address a wide range of potential ROI dimensions, we employed regional proposal anchors of various sizes ([4, 8, 16, 32, 64]), while anchor ratios ([0.5, 1.0, 2.0]) were strategically selected to accommodate different shapes of potential ROIs. 

\begin{figure}
\includegraphics[width=\textwidth]{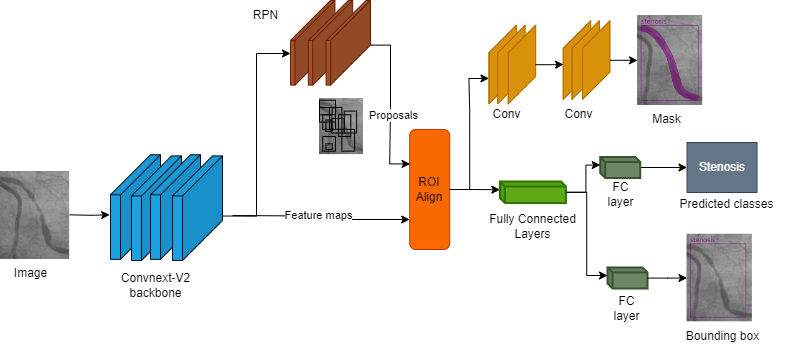}
\caption{The workflow of stenosis detection model with ConvNeXtV2 as the backbone.} \label{fig3}
\end{figure}

During the inference stage, the trained network made predictions on stenosis along with a confidence value, which indicated the likelihood of the prediction being correct. For post-processing, we used a threshold value of 0.95 on NMS for IoU threshold of RCNN and threshold of 0.8 on the confidence values of each predicted masks to generate more accurate stenosis segmentation masks. This thresholding was based on our observation that smaller thresholds led towards a greater number of false positive detections.

\subsection{Loss Functions}
During training, every training Region of Interest (RoI) is labelled with both a ground-truth class label, a target for bounding-box regression and a target mask to refine its position and  we define a multi-task loss on each sampled RoI as sum of classification loss L\textsubscript{cls}, box loss L\textsubscript{box} and mask loss L\textsubscript{mask}.

\begin{equation}
L =  \lambda_c.L\textsubscript{cls} + \lambda_b.L\textsubscript{box} + \lambda_m.L\textsubscript{mask}
\end{equation}

We use Cross-entropy loss for classification L\textsubscript{cls}, Averaged Binary Cross-entropy loss for mask loss L\textsubscript{mask} while for bounding box loss  L\textsubscript{box} we use L1 loss. The loss gain coefficients($\lambda$) are hyperparameters selected after a series of experiments on the validation set.

For true class \(y_i\) and predicted class probability \(\hat{y}_i\), classification loss L\textsubscript{cls} is defined as:
\begin{equation}
L\textsubscript{cls}(y, \hat{y}) = -\sum_{i=1}^{n} y_i \log(\hat{y}_i)
\end{equation}

For a predicted bounding box with coordinates (x, y, w, h) and a ground-truth bounding box with coordinates (x', y', w', h'), the box loss  L\textsubscript{box} is computed as:
\begin{equation}
L_{{box}}(x, y, w, h) = \sum (L1(x, x') + L1(y, y') + L1(w, w') + L1(h, h'))  
\end{equation}

For true class \(y_i\) and predicted class probability \(\hat{y}_i\) for N masks, the mask loss L\textsubscript{mask} is averaged over N masks and is given as:
\begin{equation}
L\textsubscript{mask}(y, \hat{y}) = -\frac{1}{N} \sum_{i=1}^{N} \left(y_i \log(\hat{y_i}) + (1 - y_i) \log(1 - \hat{y_i})\right) 
\end{equation}


The mask loss, L\textsubscript{mask}, is defined only on positive Region of Interests (RoIs). An RoI is considered positive if its Intersection over Union (IoU) with a ground-truth box is greater than or equal to 0.5. The mask target is obtained by taking the intersection of the RoI with its corresponding ground-truth mask.

\section{Experiments}
\subsection{Dataset, Preprocessing, and Baselines}
The ARCADE dataset~\cite{Arcadedataset} consists of 1200 images in total for each task. The spatial size of the images in the dataset is 512 x 512 pixels. In  the first phase, we split up the dataset from Phase 1 into training set with 800 images and validation set with 200 images for stenosis detection task. For the second  phase however 1000 images were used for training and 200 for validation for the models. The baseline models YOLOv8, Rtmdet, ResNet50, ResNet101 Mask R CNN and ConvNeXt backed Mask R CNN were also trained on the same 1000 training images with 200 validation set images.
To obtain the best result of stenosis detection task backed by the evidence from previous comparisons, the training set was constructed by combining training dataset with validation dataset and randomly sampling out training set of 1190 images and validation set of 10 images. 

In this study, we conducted comprehensive training on a range of baseline models, including YOLO-V8, Rtmdet-ins-large (Lyu et al., 2022)\cite{lyu2022rtmdet}, Resnet-50 Mask R-CNN, Resnet-101 Mask R-CNN, Convnext-Base Mask R-CNN, and Convnext-V2-Base Mask R-CNN. Strikingly, our findings unequivocally demonstrated the superior performance of the Convnext-V2 model compared to its counterparts. This outcome underscores the potential of Convnext-V2 as a highly effective choice for detection and segmentation tasks in the domain of medical imaging as well.

\subsection{Implementation details}
We used Convnext-V2 backbone based Mask RCNN model for stenosis detection. This architecture was trained on NVIDIA RTX 3090 graphic card for 36 epochs using AdamW optimizers ($\beta1= 0.9$, $\beta2 = 0.999$) with an initial learning rate of 1 × $10^{-4}$ and a decay rate of 0.05 per epoch with batch size of 8. For Non-Max Suppression, we kept the IOU-threshold for RPN as 0.7 for both training and inference but for RCNN we used the IoU-threshold value of 0.5 for training and 0.95 for inference after cross-validating the IoU-threshold over the range of 0.5-0.95. We selected the  IoU threshold of 0.95 along with confidence score threshold of 0.8 because higher threshold on IoU for NMS of RCNN gave better stenosis detections and higher threshold for confidence score reduced the amount of false positives. The higher value of confidence threshold ensured that the detected objects were accurately localized.

The loss function gains for box loss, class loss and mask loss were all set to 1 after some experiments conducted with the validation set. Changes in the gain coefficients did not have much effect on the performance of the models. During postprocessing even with a relatively high threshold  we encountered  multiple false positives and thus suppressed the maximum number of possible detections to 3. A series of augmentations including Random Resize, Random Crop and Random Flipping were employed during training. We also initialized the weight of our model from the pretrained model trained on MS COCO dataset\cite{lin2014microsoft} available in mmdetection library instead of training the model from scratch. We go into more details about the weight initialization in in ablation studies.

\subsection{Quantitative Evaluations}

\begin{table}
    \centering
    \begin{tabular}{|l| l|} \hline  
         \textbf{Architecture}& \textbf{F1-score}($\uparrow$)\\ \hline \hline 
         YOLO V8& 0.2318\\ 
         Rtmdet-ins-large& 0.2653\\ 
         Resnet-50 Mask R-CNN& 0.3811\\ 
         Resnet-101 Mask R-CNN& 0.4170\\  
         Convnext-Base Mask R-CNN& 0.5064\\  
         Convnext-V2-Base Mask R-CNN&\textbf{0.5353}\\ \hline
    \end{tabular}
    \captionsetup{skip=5pt}
    \caption{Comparison of F1 score on testset of different architectures on ARCADE stenosis detection task.}
    \label{tab:f1 score}
\end{table}
\vspace{-1cm}
Quantitative results in Table 1 show that ConvnNeXt-V2 backbone based Mask R-CNN achieves best overall performance in stenosis detection task under the F1 Score metric. It is also evident that ConvNeXt-V2 backbone with Mask R CNN is greatly superior when compared to traditional backbones such as ResNet50 or ResNet101, whereas it is better by than more than two times under the same metric in our dataset when compared to Yolov8 and Rtmdet. These findings also reflect on the effectiveness of the use of backbone trained using self-supervised(ConvNeXt, ConvNeXtv2) learning in medical image segmentation tasks.

\subsection{Qualitative Results}


\begin{figure}[t]
    \centering
    \begin{subfigure}[b]{0.15\textwidth}
        \centering
        \caption*{GT}
        \includegraphics[width=2cm,height=2cm]{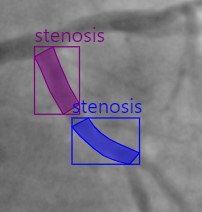}
        \label{fig:gt}
    \end{subfigure}
    \hspace{0.1cm}
    \begin{subfigure}[b]{0.15\textwidth}
        \centering
        \caption*{YOLOV8}
        \includegraphics[width=2cm,height=2cm]{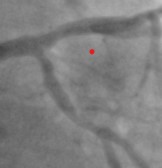}
        \label{fig:yolov8sten}
    \end{subfigure}
    \hspace{0.1cm}
    \begin{subfigure}[b]{0.15\textwidth}
        \centering
        \caption*{R101}
        \includegraphics[width=2cm,height=2cm]{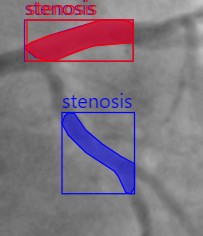}
        \label{fig:R101Mask rcnn}
    \end{subfigure}
    \hspace{0.1cm}
    \begin{subfigure}[b]{0.15\textwidth}
        \centering
        \caption*{Convnext}
        \includegraphics[width=2cm,height=2cm]{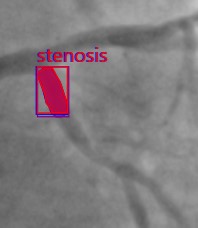}
        \label{fig:Convnextv1 mask rcnn}
    \end{subfigure}
    \hspace{0.1cm}
    \begin{subfigure}[b]{0.15\textwidth}
        \centering
        \caption*{ConvnextV2}
        \includegraphics[width=2cm,height=2cm]{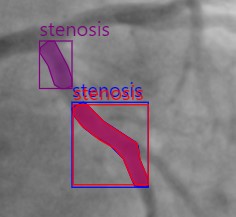}
        \label{fig:Convnext V2}
    \end{subfigure}\\
    \begin{subfigure}[b]{0.15\textwidth}
        \centering
        \includegraphics[width=2cm,height=2cm]{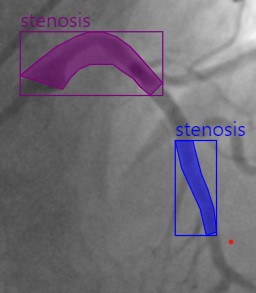}
        \label{fig:gt}
    \end{subfigure}
    \hspace{0.1cm}
    \begin{subfigure}[b]{0.15\textwidth}
        \centering
        \includegraphics[width=2cm,height=2cm]{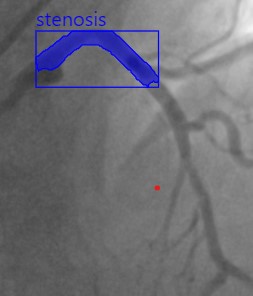}
        \label{fig:yolov8sten}
    \end{subfigure}
    \hspace{0.1cm}
    \begin{subfigure}[b]{0.15\textwidth}
        \centering
        \includegraphics[width=2cm,height=2cm]{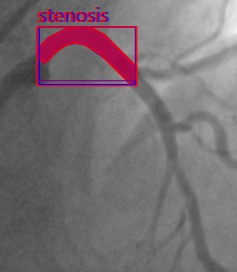}
        \label{fig:R101Mask rcnn}
    \end{subfigure}
    \hspace{0.1cm}
    \begin{subfigure}[b]{0.15\textwidth}
        \centering
        \includegraphics[width=2cm,height=2cm]{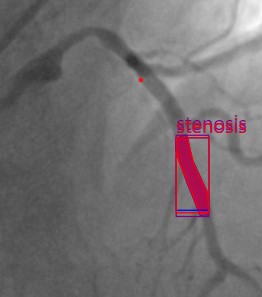}
        \label{fig:Convnextv1 mask rcnn}
    \end{subfigure}
    \hspace{0.1cm}
    \begin{subfigure}[b]{0.15\textwidth}
        \centering
        \includegraphics[width=2cm,height=2cm]{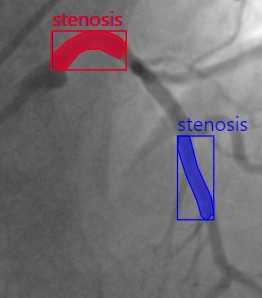}
        \label{fig:Convnext V2}
    \end{subfigure}\\

    \begin{subfigure}[b]{0.15\textwidth}
        \centering
        \includegraphics[width=2cm,height=2cm]{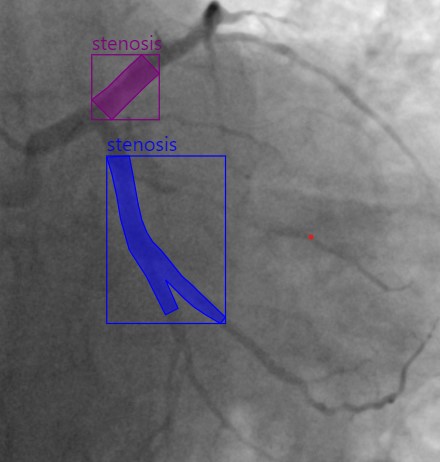}
        \label{fig:gt}
    \end{subfigure}
    \hspace{0.1cm}
    \begin{subfigure}[b]{0.15\textwidth}
        \centering
        \includegraphics[width=2cm,height=2cm]{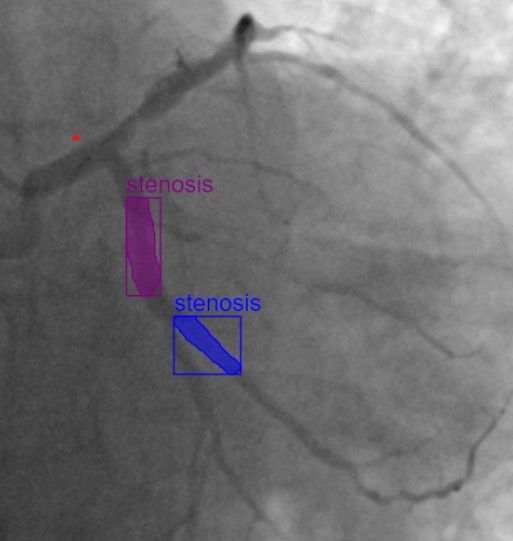}
        \label{fig:yolov8sten}
    \end{subfigure}
    \hspace{0.1cm}
    \begin{subfigure}[b]{0.15\textwidth}
        \centering
        \includegraphics[width=2cm,height=2cm]{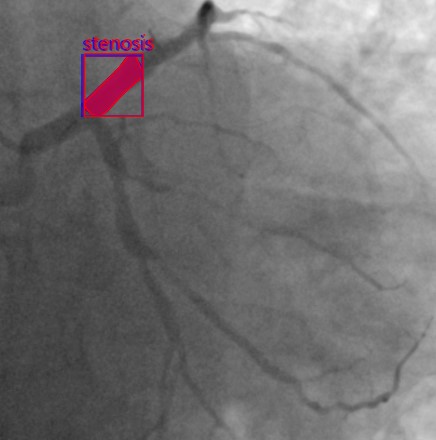}
        \label{fig:R101Mask rcnn}
    \end{subfigure}
    \hspace{0.1cm}
    \begin{subfigure}[b]{0.15\textwidth}
        \centering
        \includegraphics[width=2cm,height=2cm]{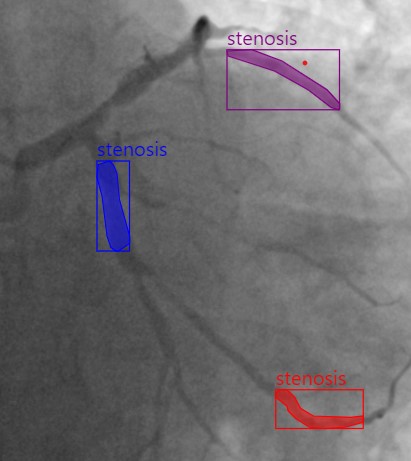}
        \label{fig:Convnextv1 mask rcnn}
    \end{subfigure}
    \hspace{0.1cm}
    \begin{subfigure}[b]{0.15\textwidth}
        \centering
        \includegraphics[width=2cm,height=2cm]{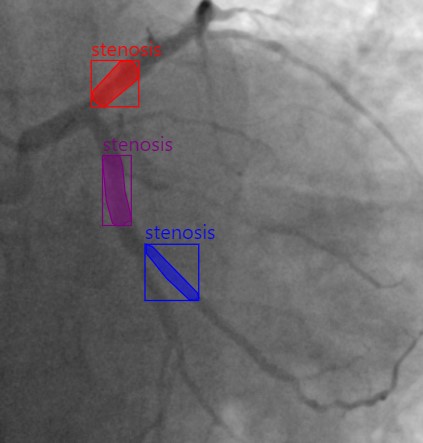}
        \label{fig:Convnext V2}
    \end{subfigure}\\
    
    \begin{subfigure}[b]{0.15\textwidth}
        \centering
        \includegraphics[width=2cm,height=2cm]{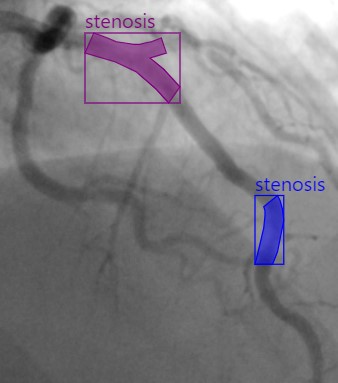}
        \label{fig:gt}
    \end{subfigure}
    \hspace{0.1cm}
    \begin{subfigure}[b]{0.15\textwidth}
        \centering
        \includegraphics[width=2cm,height=2cm]{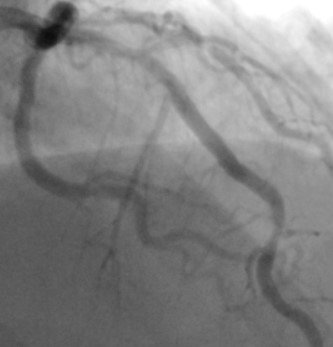}
        \label{fig:yolov8sten}
    \end{subfigure}
    \hspace{0.1cm}
    \begin{subfigure}[b]{0.15\textwidth}
        \centering
        \includegraphics[width=2cm,height=2cm]{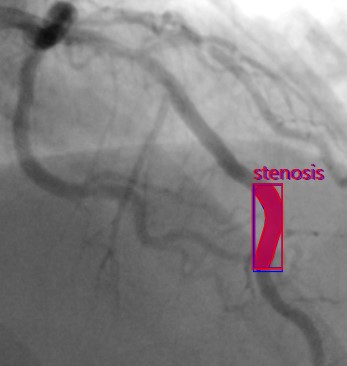}
        \label{fig:R101Mask rcnn}
    \end{subfigure}
    \hspace{0.1cm}
    \begin{subfigure}[b]{0.15\textwidth}
        \centering
        \includegraphics[width=2cm,height=2cm]{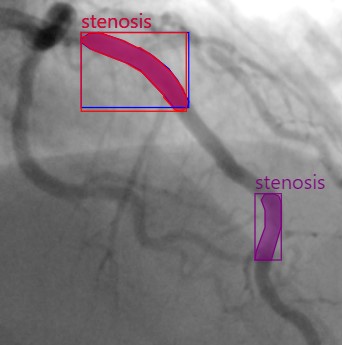}
        \label{fig:Convnextv1 mask rcnn}
    \end{subfigure}
    \hspace{0.1cm}
    \begin{subfigure}[b]{0.15\textwidth}
        \centering
        \includegraphics[width=2cm,height=2cm]{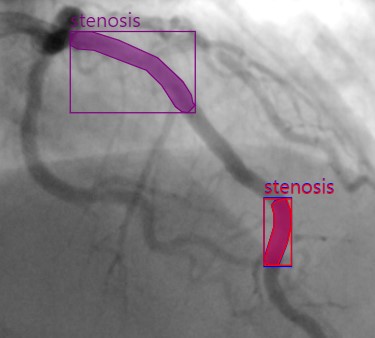}
        \label{fig:Convnext V2}
    \end{subfigure}
    \captionsetup{skip=5pt}
    \caption{Qualitative instance segmentation results on stenosis detection. Ground truth masks followed by the instance segmentation masks generated by Yolo-V8, Resnet101 Mask R-CNN, Convnext Mask R-CNN and Convnext-V2 Mask R-CNN are shown in the figure respectively.}
    \label{fig:Stenosis Qualitative Comparison}
\end{figure}

Figure 2 shows qualitative segmentation results on unseen images, the test  dataset. We see that our model can accurately detect and segment the structure of coronary arteries and correctly identify the location of stenosis under different circumstances and view angles. This comparison sheds light to the fact that ConvNeXtv2 indeed greatly enhances the segmentation capability of the Mask R-CNN architecture while the other backbones using the same segmentation method generate subpar predictions even on images with reasonable contrast and illumination.

\subsection{Ablation Studies}

We conducted an extensive evaluation of weight initialization methods for our instance segmentation model. Specifically, we compared the performance of models initialized with pretrained weights from the MS-COCO dataset against those initialized using Xavier Initialization, as outlined in Glorot et al. ~\cite{pmlr-v9-glorot10a}. Our results in Table 2 unequivocally demonstrates that leveraging pretrained weights from model trained on the MS-COCO dataset yields superior performance compared to Xavier Initialization. This is mainly due to the fact that the model pretrained on segmentation task has already learned the features important for segmentation and focuses on a much specialized task when finetuned on ARCADE dataset.

\begin{table}[ht]
  \begin{minipage}{0.5\textwidth}
    \centering
    \begin{tabular}{|c|c|}
      \hline
      \textbf{Weight Initialization} & \textbf{F1-score($\uparrow$)} \\
      \hline
      Xavier Initialization & 0.43 \\
      Pretrained on MS-COCO & \textbf{0.53} \\
      \hline
    \end{tabular}
    \captionsetup{skip=5pt}
    \caption{Comparison of Initialization}
  \end{minipage}%
  \begin{minipage}{0.5\textwidth}
    \centering
    \begin{tabular}{|c|c|}
      \hline
      \textbf{Architecture}& \textbf{seg-MAP}($\uparrow$)\\
      \hline
      Multi-Task & 0.109 \\
      Single task(stenosis)&\textbf{0.186} \\
      \hline
    \end{tabular}
    \captionsetup{skip=5pt}
    \caption{Seg-MAP on validation}
  \end{minipage}
\end{table}

\definecolor{bblue}{HTML}{4F81BD}

\begin{figure}
    \centering
\begin{tikzpicture}
    \begin{axis}[
        width  = \textwidth,
        height = 5cm,
        major x tick style = transparent,
        ybar=2*\pgflinewidth,
        bar width=14pt,
        ymajorgrids = true,
        ylabel = {F1-Score},
        xlabel = {IoU-threshold},
        xtick = {0.5,0.55,0.6,0.65,0.7,0.75,0.8,0.85,0.9,0.95},
        ymin=0.4,
        ymax=0.6,
        scaled y ticks = false,
        enlarge x limits=0.05,
    ]
        \addplot[style={bblue,fill=bblue,mark=none}]
            coordinates {(0.5, 0.4605) (0.55, 0.4608) (0.6, 0.4610) (0.65, 0.4621) (0.7, 0.4639) (0.75, 0.4677) (0.8,0.4803)(0.85,0.4960)(0.9,0.5211)(0.95,0.5353)}
            node[pos=0,yshift=2mm] {0.4605}
            node[pos=0.1,yshift=2mm] {0.4608}
            node[pos=0.2,yshift=2mm] {0.4610}
            node[pos=0.3,yshift=2mm] {0.4621}
            node[pos=0.4,yshift=2mm] {0.4639}
            node[pos=0.5,yshift=2mm] {0.4677}
            node[pos=0.7,yshift=2mm] {0.4803}
            node[pos=0.8,yshift=2mm] {0.4960}
            node[pos=0.9,yshift=2mm] {0.5211}
            node[pos=1,yshift=2mm] {0.5353};
    \end{axis}
\end{tikzpicture}

    \captionsetup{skip=5pt}
    \caption{Comparison of F1 score for different IoU threshold value of RCNN's NMS for Convnext-V2 Mask RCNN Architecture on testset of ARCADE stenosis detection task.}
\end{figure}
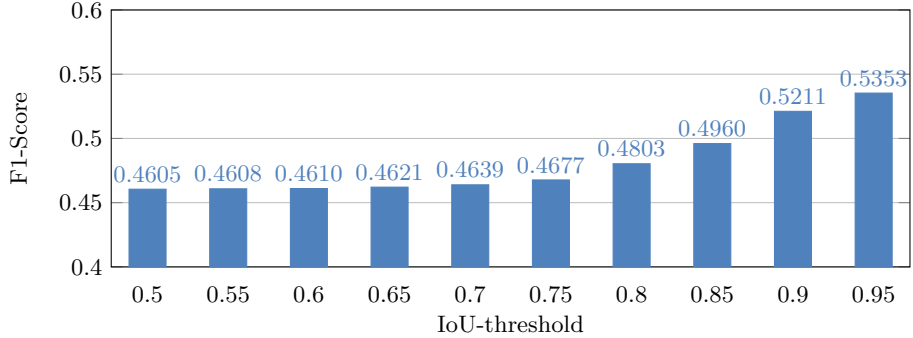

Furthermore, we tried to enforce learning of both vessel segmentation task and stenosis detection task of ARCADE dataset using same model. We structured model to include separate prediction heads for seperate tasks using a shared base model, while still fine-tuning its predictions specifically for each task. But, the model failed to learn any useful features after few epochs eventually saturating on a quite low seg-MAP score. Table 3 showcases that model trained on only stenosis task achieves far better performance than multitask model trained on both tasks.

Additionally, we undertook a thorough assessment of the Convnext-V2 Mask RCNN model's performance across various IOU thresholds. Figure 3 describes the model and its relation with IOU thresholds.Our model achieves its optimal performance at an IOU threshold of 0.95 when employing the Non-Max Suppression Algorithm of RCNN. 

    


\section{Conclusion}

This paper presents an innovative deep learning method for instance segmentation of coronary arteries. The proposed framework contains two key components, the use of Convnext-V2 in Mask-RCNN framework and confidence score threshold based post processing. The outcome can benefit from the more effective feature maps from Convnext-V2 backbone and confidence score thresholding results in reduction of false positives in instance segmentation. Extensive experiments are conducted on ARCADE dataset. The results suggest that Convnext-V2 backbone based Mask R-CNN model can achieve competitive performance with the state-of-the-art instance segmentation models.

%
%
%
\bibliographystyle{splncs04.bst}
\bibliography{reference.bib}





\end{document}